\documentclass[final]{dynafront2026}

\usepackage{booktabs}

\title[Uncertainty-Guided Multimodal Diffusion]
{Uncertainty-Guided Cross-Modal Alignment for Diffusion-Based IoT Intrusion Detection}

\optauthor{%
  \Name{Mahdi Arab Loodaricheh}, 
  \Name{Anita Raja}, and
  \Name{Mohammad Hossein Manshaei}\\[-0.35em]
  \Email{mahdi.arabloodaricheh41@gc.cuny.edu;
  \{anita.raja,mohammad.manshaei\}@hunter.cuny.edu}\\[-0.35em]
  \addr Department of Computer Science, Hunter College and The Graduate Center,
  CUNY, New York, NY, USA
}

\hypersetup{
  pdftitle={Uncertainty-Guided Cross-Modal Alignment for Diffusion-Based IoT Intrusion Detection},
  pdfauthor={Mahdi Arab Loodaricheh, Anita Raja, Mohammad Hossein Manshaei},
  pdfsubject={DynaFront 2026 workshop paper},
  pdfkeywords={conditional diffusion, multimodal learning, uncertainty, intrusion detection}
}

\begin{document}
\maketitle

\begin{abstract}
Diffusion models evolve noisy states toward a learned data distribution, so
their sampling behavior depends on the geometry of the representation in
which denoising occurs.  This dependence is especially important for
multimodal data: independently trained encoders can map synchronized views
to incompatible latent regions, while uniform alignment can let an
uncertain view distort the shared geometry. To address this gap, we introduce a novel reliability-guided alignment rule
that uses normalized VAE posterior variance to control directional cross-modal influence before diffusion training. We study this issue in
minority-attack generation for imbalanced Internet-of-Things intrusion
detection.  Tabular flow attributes, a DeepInsight image, and a packet
sequence are encoded by modality-specific variational autoencoders.
Per-modality normalized VAE posterior variances provide bounded,
sample-specific weighting signals in an asymmetric stop-gradient alignment
objective. We use these quantities as heuristic encoder-derived reliability
proxies rather than calibrated probabilistic uncertainties. The full
modality-specific latent means obtained after alignment fine-tuning are
standardized and modeled jointly by a class-conditional EDM; a second-order
sampler then generates paired attack-conditioned latent blocks for the three
decoders.
In controlled experiments on TON-IoT, uncertainty-aware alignment obtains
the highest mean precision, recall, density, and coverage among three
diffusion variants.  Augmenting only the training split yields ROC-AUC
$0.9515$, PR-AUC $0.8954$, balanced accuracy $0.8395$, and attack-class F1
$0.7940$ on an untouched real test set.  Among the two variants with matched alignment fine-tuning, posterior-variance-based weighting improves both tabular generative fidelity and downstream detection relative to uniform weighting.
\end{abstract}


\section{Introduction}
\label{sec:introduction}

Diffusion-based generation transports a simple noisy distribution toward a
structured data distribution.  In latent diffusion, this transport inherits
its geometry from the encoders: latent scales and correspondences determine
what the denoiser must learn and how a numerical sampler traverses the state
space.  Tabular diffusion benefits from compact latent representations
\citep{Kotelnikov2023TabDDPM,Zhang2024TabSyn}, while MAGE-ID couples tabular
flows and image transformations through a joint latent diffusion prior
\citep{ArabLoodaricheh2026MAGEID}.  Adding packet sequences creates a
three-view constraint: every generated block must describe the same event.

In imbalanced IoT intrusion detection, random oversampling adds no
observations and SMOTE is limited to local interpolation
\citep{Chawla2002SMOTE}, whereas conditional generation can model the
minority distribution.

Independent encoders do not guarantee synchronization, whereas uniform
alignment assumes every view is equally informative for every sample.  A
noisy packet sequence or ambiguous image can then pull reliable
representations toward an uncertain target, compounding the modality
imbalance observed in multimodal VAEs~\citep{Javaloy2022ModalityCollapse}.
Reliability weighting is useful in multimodal fusion
\citep{Zhang2023QMF,Gao2024EAU}; here it shapes the state space on which
conditional denoising is learned.

In our method, modality-specific VAEs first encode the three views. The novel technical component is a bounded, sample-dependent asymmetric
alignment objective in which each target modality's normalized VAE
posterior-variance proxy controls its influence on the other encoders. We use
mean VAE posterior variance as a heuristic, encoder-derived reliability proxy
and normalize it separately for each modality. The resulting bounded,
sample-specific weights control an asymmetric stop-gradient alignment
objective. The full modality-specific latent means are then standardized,
concatenated, and modeled by a class-conditional EDM
\citep{karras2022elucidating}. The uniform and posterior-variance-weighted
variants receive identical alignment fine-tuning and differ only in their
target weights; their comparison therefore evaluates the effect of the
weighting rule.

\section{Reliability-Aware Joint Latent Dynamics}
\label{sec:method}

\subsection{Multimodal state construction}

Let $\mathbf{x}_n=(\mathbf{x}^{t}_n,\mathbf{x}^{i}_n,
\mathbf{x}^{s}_n,y_n)$ contain tabular, image, and sequence views and a
normal/attack label.  We learn their joint distribution conditioned on
$y=1$ and add generated attacks only to training.  The 22 flow features are
also mapped to an $8\times8$ DeepInsight image
\citep{sharma2019deepinsight}; the sequence retains 50 packets with eight
attributes.  Common indices preserve correspondence.  Transformer,
convolutional, and LSTM VAEs encode the views.  Each diagonal-Gaussian
posterior has mean $\boldsymbol{\mu}^{m}$ and log-variance
$\boldsymbol{\ell}^{m}$ and is pretrained with reconstruction and KL losses
\citep{kingma2014autoencoding}; its full mean has dimension 92, 32, or 16.

\begin{figure}[t]
  \centering
  \includegraphics[width=\textwidth]{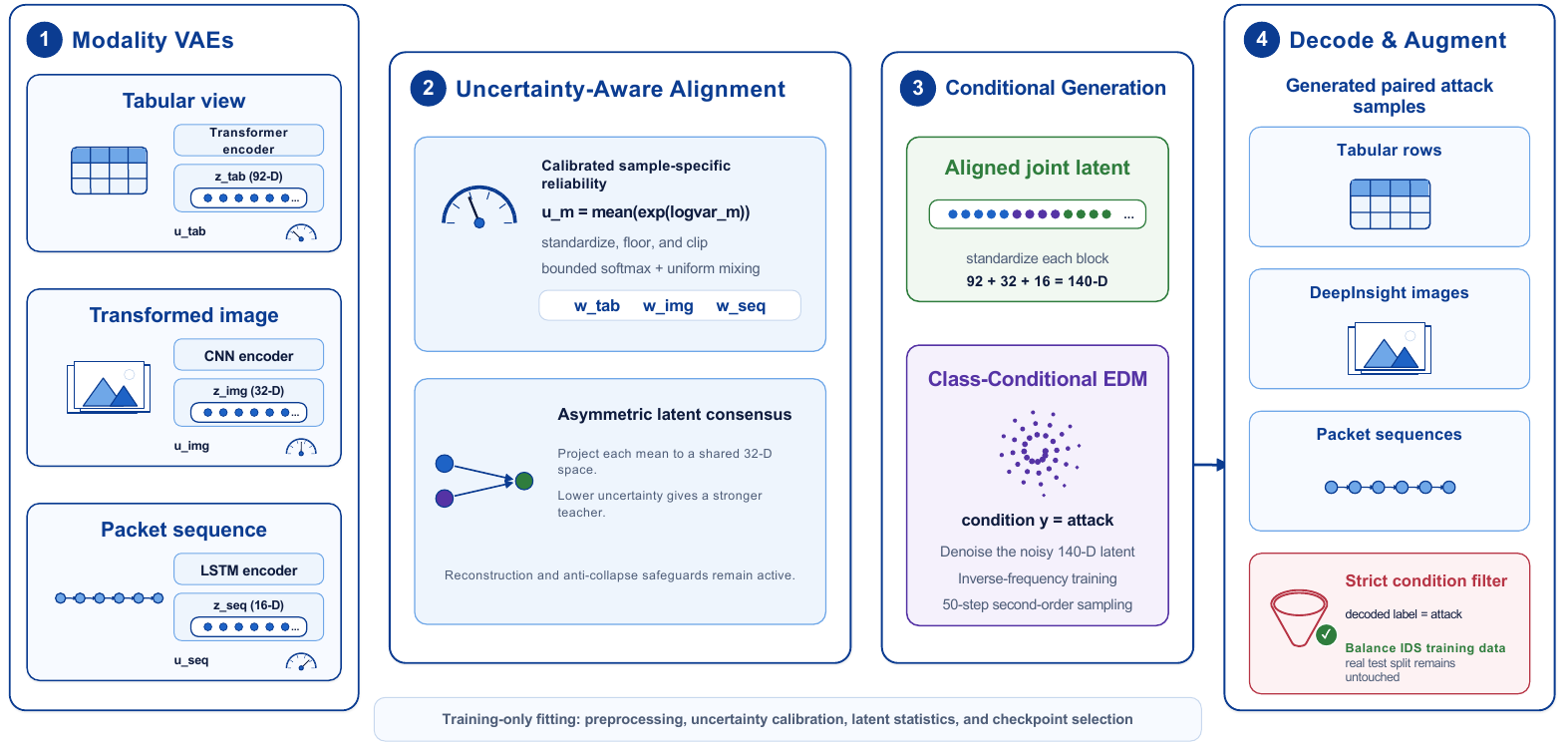}
  \caption{The proposed pipeline.  Reliability-weighted alignment shapes
  an auxiliary shared geometry; class-conditional EDM operates on the
  standardized full latent blocks and decodes synchronized attack views.
  All fitted transformations and checkpoint decisions use training data
  only.}
  \label{fig:framework}
\end{figure}

\subsection{Uncertainty-weighted alignment}

For sample $n$, modality $m$ has mean posterior variance
$u_{n,m}=d_m^{-1}\sum_r\exp(\ell^{m}_{n,r})$.  Because encoder scales
differ, we standardize $\log u_{n,m}$ using per-modality statistics from a
fixed clean training subset, apply a deviation floor, and clip the result to
$\widetilde u_{n,m}$.  Confidence and the bounded weight are
\begin{equation}
a_{n,m}=\frac{\exp(-\gamma\widetilde u_{n,m})}
{\sum_k\exp(-\gamma\widetilde u_{n,k})},\qquad
w_{n,m}=\eta a_{n,m}+(1-\eta)/3,
\label{eq:weight}
\end{equation}
with $\gamma=2$ and $\eta=0.5$.  Thus, confidence changes relative
influence without assigning zero weight to any view.  The weights are
detached so an encoder cannot reduce the objective by manipulating its
reported variance.

Projection heads $P_m$ map posterior means to a shared 32-dimensional space,
$\mathbf{h}^{m}_n=P_m(\boldsymbol{\mu}^{m}_n)$.  For mini-batch
$\mathcal{B}$, alignment uses
\begin{equation}
\mathcal{L}_{\mathrm{align}}=
\frac{1}{6|\mathcal{B}|}
\sum_{n\in\mathcal{B}}\sum_m\sum_{k\ne m}
w_{n,k}\left\|\mathbf{h}^{m}_n-
\operatorname{sg}(\mathbf{h}^{k}_n)\right\|_2^2,
\label{eq:alignment}
\end{equation}
where $\operatorname{sg}$ stops gradients through the target.  The target's
weight, rather than the source's, controls the update: a lower-uncertainty
view is a stronger teacher.  Reconstruction and KL losses remain active,
and a variance-floor penalty discourages collapsed projection dimensions.
The projection heads impose correspondence but are not passed to the
diffusion model.

\subsection{Conditional denoising trajectory}

The complete means are standardized blockwise with training statistics and
concatenated into $\overline{\mathbf{z}}\in\mathbb{R}^{140}$.  This removes
coordinate-scale differences that could otherwise let one modality dominate
the denoising loss.  EDM samples a noise level $\sigma$, forms
$\widetilde{\mathbf{z}}=\overline{\mathbf{z}}+
\sigma\boldsymbol{\epsilon}$, and minimizes
\begin{equation}
\mathcal{L}_{\mathrm{diff}}=
\mathbb{E}\!\left[\omega(\sigma)
\left\|D_\theta(\widetilde{\mathbf{z}},\sigma,y)
-\overline{\mathbf{z}}\right\|_2^2\right].
\label{eq:diffusion}
\end{equation}
An inverse-frequency sampler balances exposure to the two classes during
training.  With $\sigma$ as the integration variable, the EDM derivative is
$\mathrm{d}\widetilde{\mathbf{z}}/\mathrm{d}\sigma=
(\widetilde{\mathbf{z}}-
D_\theta(\widetilde{\mathbf{z}},\sigma,y))/\sigma$.
Starting from Gaussian noise and fixing $y=1$, 50 second-order updates
follow this direction toward an attack-conditioned latent.  The state is
split, returned to each block's original scale, and decoded.  A candidate is
retained only when the reconstructed categorical label agrees with the
requested attack condition.

Alignment changes the representation geometry through directed,
reliability-weighted updates; the EDM sampler then transports noise through
that geometry.  Alignment makes corresponding blocks compatible, while
uncertainty weighting limits distortion from an unreliable target.

\section{Existing Experimental Study}
\label{sec:experiments}

\subsection{Protocol}

Experiments use the network-traffic component of TON-IoT
\citep{Alsaedi2020TONIoT}.  The binary target combines every non-normal
attack category while retaining original multiclass labels during
stratification.  The downstream study uses a fixed 30,000-event, 80:20 pool
split into 21,000 training, 3,000 validation, and 6,000 test observations;
training contains 16,800 normal and 4,200 attack events.  PRDC uses
independently prepared 30,000-event artifacts under the same class ratio,
split proportions, and saved-index protocol.  Comparisons within each
evaluation therefore use identical observations.  Preprocessing, image
coordinates, normalization, latent statistics, uncertainty normalization,
and checkpoint selection are fitted without test data.

We compare real-only, class-weighted, random-oversampled, and SMOTE training
with three diffusion variants.  No alignment uses the
independently pretrained VAE latents; uniform alignment sets every modality
weight to $1/3$; uncertainty-aware alignment uses Eq.~\eqref{eq:weight}.
The three diffusion variants share encoders, latent dimensions, denoiser,
optimizer settings, class condition, sampler, augmentation volume, and data
split.  Each augments the training set to equal class counts.  Downstream
utility is measured by one fixed XGBoost configuration
\citep{Chen2016XGBoost}; the test set remains real and imbalanced.  We report
mean and sample standard deviation over three model seeds.  Precision,
recall, density, and coverage (PRDC) compare equal-sized sets of real and
generated tabular attacks~\citep{Naeem2020PRDC}.

The VAEs train for at most 100 epochs, followed by 50 alignment epochs.  The
alignment coefficient reaches $0.01$ after a linear warm-up; the
anti-collapse penalty has weight $10^{-3}$.  Log-uncertainties use a $0.05$
standard-deviation floor and are clipped to $[-5,5]$.  The diffusion model
trains for at most 1,000 epochs with early-stopping patience 200.  NumPy,
PyTorch, synthetic-sample selection, and XGBoost use three model seeds while
the saved data split is held fixed.

\begin{table}[t]
\centering
\caption{PRDC for generated tabular attacks (mean $\pm$ standard deviation
over three seeds).  Higher is better.}
\label{tab:prdc}
\footnotesize
\begin{tabular*}{\textwidth}{@{\extracolsep{\fill}}lcccc@{}}
\toprule
Method & Precision & Recall & Density & Coverage \\
\midrule
No alignment
 & $0.6688\pm0.1015$ & $0.8563\pm0.0775$
 & $0.6616\pm0.0969$ & $0.7093\pm0.0897$ \\
Uniform alignment
 & $0.7248\pm0.0137$ & $0.9054\pm0.0181$
 & $0.7090\pm0.0154$ & $0.7696\pm0.0063$ \\
\textbf{Uncertainty-aware}
 & $\mathbf{0.7341\pm0.0316}$ & $\mathbf{0.9067\pm0.0194}$
 & $\mathbf{0.7229\pm0.0403}$ & $\mathbf{0.7921\pm0.0185}$ \\
\bottomrule
\end{tabular*}
\end{table}

\subsection{Fidelity and downstream utility}

Table~\ref{tab:prdc} isolates the geometry used by the same conditional
diffusion process.  Uniform alignment improves every mean relative to no
alignment.  Its coverage standard deviation falls from $0.0897$ to $0.0063$.
The uncertainty-aware variant obtains the highest mean in all four columns;
coverage rises from $0.7696$ to $0.7921$ relative to uniform alignment while
the other means do not decrease.  Its standard deviations are higher than
the uniform variant's for three metrics, so the result supports better mean
fidelity rather than uniformly lower variance.

\begin{table}[t]
\centering
\caption{Downstream performance on the untouched real test set (mean
$\pm$ standard deviation over three seeds).  BAcc. denotes balanced
accuracy; higher is better.}
\label{tab:downstream}
\footnotesize
\begin{tabular*}{\textwidth}{@{\extracolsep{\fill}}lcccc@{}}
\toprule
Training strategy & ROC-AUC & PR-AUC & BAcc. & F1 \\
\midrule
Real only
 & $0.9410\pm0.0014$ & $0.8799\pm0.0013$
 & $0.8244\pm0.0018$ & $0.7821\pm0.0029$ \\
Class weighted
 & $0.9418\pm0.0015$ & $0.8786\pm0.0014$
 & $0.8212\pm0.0016$ & $0.7213\pm0.0051$ \\
Random oversampling
 & $0.9416\pm0.0018$ & $0.8783\pm0.0021$
 & $0.8207\pm0.0027$ & $0.7209\pm0.0048$ \\
SMOTE
 & $0.9376\pm0.0018$ & $0.8716\pm0.0033$
 & $0.8310\pm0.0026$ & $0.7349\pm0.0086$ \\
No-align. diffusion
 & $0.9381\pm0.0022$ & $0.8747\pm0.0025$
 & $0.8243\pm0.0025$ & $0.7800\pm0.0035$ \\
Uniform-align. diffusion
 & $0.9386\pm0.0008$ & $0.8750\pm0.0011$
 & $0.8245\pm0.0019$ & $0.7797\pm0.0022$ \\
\textbf{Uncertainty-aware}
 & $\mathbf{0.9515\pm0.0003}$ & $\mathbf{0.8954\pm0.0006}$
 & $\mathbf{0.8395\pm0.0024}$ & $\mathbf{0.7940\pm0.0022}$ \\
\bottomrule
\end{tabular*}
\end{table}

Table~\ref{tab:downstream} shows that class balance alone is insufficient.
Class weighting and random oversampling reduce balanced accuracy and F1
relative to real-only training; SMOTE improves balanced accuracy but not the
other three measures.  No-alignment and uniform-alignment diffusion are also
nearly indistinguishable downstream.  In contrast, uncertainty-aware
alignment has the best mean in every column.  Against uniform alignment it
adds $0.0129$ ROC-AUC, $0.0204$ PR-AUC, $0.0150$ balanced accuracy, and
$0.0143$ F1.  Against the strongest alternative in each column, the gains
are $0.0097$, $0.0155$, $0.0085$, and $0.0119$, respectively.

Uniform alignment improves fidelity but not the classifier.  Reliability-
aware alignment improves both under the same denoising and sampling setup,
consistent with the hypothesis that a shared geometry is insufficient when
uncertain views remain equally strong targets.  With only three seeds, we
treat the differences as controlled evidence, not statistical significance.

\section{Discussion}
\label{sec:discussion}

The results distinguish two roles of multimodal latent alignment. Uniform
alignment improves all PRDC means relative to independent latent modeling,
suggesting that cross-modal compatibility provides a more coherent state
space for conditional denoising. However, this improvement does not translate
into comparable downstream gains. Generative fidelity alone is therefore
insufficient: the aligned representation must also preserve minority-class
structure that is useful to the detector. Uncertainty-aware alignment improves
both fidelity and downstream performance, supporting the use of selective,
sample-dependent cross-modal influence rather than treating every modality as
an equally reliable target.

This interpretation also connects representation learning to the resulting
denoising dynamics. Blockwise standardization prevents latent scale differences
from dominating the EDM objective, bounded reliability weights preserve all
modalities while varying their influence, and stop-gradient targets make the
direction of alignment explicit. Together, these mechanisms shape the geometry
through which the sampler transports noise toward an attack-conditioned
multimodal state. The present endpoint measurements motivate future analysis
of the complete trajectory, including path length, curvature, solver error,
and cross-modal agreement throughout sampling.
\paragraph{Future Work.}
Future work will extend the evaluation to multiclass and cross-dataset
settings, examine joint multimodal fidelity, and investigate more formal
measures of modality uncertainty and denoising stability.

\section{Conclusion}

Our work shows that calibrated reliability can guide multimodal alignment
while bounded weights and anti-collapse safeguards preserve every view. The
conditional EDM then models the resulting joint geometry to generate
synchronized attack latents. Under the same denoising and sampling setup,
uniform alignment improves generative fidelity but provides little downstream
benefit, whereas uncertainty-aware alignment improves both fidelity and
intrusion-detection performance. These results demonstrate that effective
multimodal diffusion depends not only on aligning representations, but also
on controlling which modalities are trusted during alignment.

\clearpage
\bibliography{references}

\end{document}